\documentclass{svproc}
\usepackage{url}

\usepackage{booktabs}
\usepackage{graphicx} 
\usepackage{float}
\usepackage{booktabs} 
\usepackage{array} 
\usepackage{multirow}
\usepackage[table,xcdraw]{xcolor} 
\begin{document}
\mainmatter              
\title{ Machine Learning-Based Prediction of ICU Readmissions in Intracerebral Hemorrhage Patients: Insights from the MIMIC Databases}
\titlerunning{ML-based Prediction of ICU
Readmissions in ICH Patients}  
%
\author{Shuheng Chen\inst{1} \and Junyi Fan\inst{1} \and Armin Abdollahi\inst{1} \and Negin Ashrafi\inst{1} \and Kamiar Alaei\inst{2} \and Greg Placencia\inst{3} \and Maryam Pishgar\inst{1}}
\authorrunning{Chen et al.} 
%
\tocauthor{Shuheng Chen, Junyi Fan, Armin Abdollahi, Negin Ashrafi, Kamiar Alaei, Greg Placencia and Maryam Pishgar}
\institute{
University of Southern California, Los Angeles, USA\inst{1} \and
California State University, Long Beach, USA\inst{2} \and
California State Polytechnic University, Pomona, USA\inst{3} \newline
\email{pishgar@usc.edu}
}

\maketitle              

\begin{abstract} 

Intracerebral hemorrhage (ICH) is a life-risking condition characterized by bleeding within the brain parenchyma. ICU readmission in ICH patients is a critical outcome, reflecting both clinical severity and resource utilization. Accurate prediction of ICU readmission risk is crucial for guiding clinical decision-making and optimizing healthcare resources.This study utilized the Medical Information Mart for Intensive Care (MIMIC-III and MIMIC-IV) databases, which contain comprehensive clinical and demographic data on ICU patients. Patients with ICH were identified from both databases. Various clinical, laboratory, and demographic features were extracted for analysis based on both overview literature and experts’ opinions. Preprocessing methods like imputing and sampling were applied to improve the performance of our models. Machine learning techniques, such as Artificial Neural Network (ANN), XGBoost, and RandomForest were employed to develop predictive models for ICU readmission risk. Model performance was evaluated using metrics such as AUROC, accuracy, sensitivity, and specificity.The developed models demonstrated robust predictive accuracy for ICU readmission in ICH patients, with key predictors including demographic information, clinical parameters, and laboratory measurements. Our study provides a predictive framework for ICU readmission risk in ICH patients, which can aid in clinical decision-making and improve resource allocation in intensive care settings.
\end{abstract}
\section{Background}
Intracerebral hemorrhage (ICH) is a severe form of stroke caused by bleeding into the brain parenchyma \cite{magid2022cerebral}. It accounts for approximately 10\% of all strokes in the USA \cite{benjamin2019american} and 6.5\%–19.6\% globally \cite{feigin2017global}. ICH has a high mortality rate, with up to 50\% of patients dying within 30 days of diagnosis \cite{van2010incidence,pinho2019intracerebral}. In low- and middle-income countries, its higher incidence contributes to a disproportionate burden of disability-adjusted life years compared to ischemic stroke \cite{feigin2017global,globalich2020,krishnamurthi2020global}.

Managing ICH patients is particularly challenging due to the poor prognosis of the condition, rapid neurological deterioration, and high rates of morbidity and mortality. Effective management requires timely neuroimaging, blood pressure control, and, in some cases, surgical intervention, such as hematoma evacuation, which requires a multidisciplinary approach \cite{mcgurgan2021acute,kirshner2021management}. Emerging strategies, including minimally invasive techniques and predictive modeling, aim to improve outcomes and optimize resource use \cite{elias2024new}.

ICU readmission is a key metric for assessing care quality and patient stability. For ICH patients, predicting readmission risk can improve clinical outcomes, reduce costs, and optimize resource allocation. Although studies specific to ICH-related readmission are limited, research on predictive models highlights their potential to identify high-risk patients and enable timely interventions \cite{hosein2014meta,maharaj2018utility}.

Machine learning (ML) has shown transformative potential in medical data analysis, enabling the identification of complex patterns in high-dimensional datasets \cite{li2024machine}. Among the advanced ML techniques, Artificial Neural Network (ANN), XGBoost, and Random Forest have demonstrated particular effectiveness due to their ability to model non-linear relationships, handle missing data, and incorporate regularization to prevent overfitting. From these methods, ANN has garnered significant attention in recent years for its superior performance in clinical prediction tasks, such as disease classification and risk stratification, showing its significant promise in clinical prediction tasks. For instance, Esteva et al. (2020) used deep learning models based on an ANN model to diagnose skin cancer from dermatological images, achieving performance comparable to expert dermatologists \cite{esteva2019guide}. In another study, Li et al. (2021) employed ANN to predict diabetic retinopathy from retinal images, surpassing traditional diagnostic methods in accuracy and early detection \cite{tsiknakis2021deep}. Rajkomar et al. (2021) applied an ANN model to predict hospital readmissions using electronic health records (EHR), demonstrating that deep learning models outperform traditional methods like the LACE index in predictive accuracy \cite{rajkomar2018scalable}.XGBoost is also a highly efficient machine learning algorithm known for its accuracy, handling of imbalanced data, and interoperability \cite{chen2016xgboost}. In clinical predictions, it is valuable for predicting patient outcomes, such as readmission risk, by analyzing large, complex datasets, enabling early interventions and optimized care. For instance, Li et al. (2023) successfully used an XGBoost model to predict dynamic sepsis onset in ICU patients. The model was evaluated on data from ICU patients and demonstrated enhanced sensitivity and specificity, making it a promising tool for improving clinical decision-making and early intervention in critical care \cite{liu2022dynamic}.In the case of Random Forest, Zhou et al. (2022) constructed an integrated predictive model for chronic kidney disease (CKD) risk, combining both Random Forest and ANN. The study demonstrated that the hybrid model achieved superior performance over individual models in predicting CKD risk, making it a valuable tool for clinical decision-making \cite{zhou2022construction}.

This study aims to develop a machine learning model tailored to predict readmissions from the ICU in ICH patients. By incorporating comprehensive clinical and demographic data, including comorbidities and prior hospitalizations, the model aims to achieve high predictive accuracy. Research adheres to established reporting standards to ensure methodological rigor, transparency, and reproducibility.

\section{Methodology}\label{sec:Methodology}
\subsection{Data Source}\label{subsec:Data Source}
The data used in this study were extracted from the publicly available Medical Information Mart for Intensive Care (MIMIC) databases, specifically MIMIC-III \cite{johnson2016mimic} and MIMIC-IV \cite{johnson2023mimic}.These databases provide an extensive and unidentified collection of clinical information from patients in the ICU, which includes demographics, vital signs, laboratory test results, administered medications, and clinical notes. The deidentification process ensures strict adherence to patient privacy regulations while facilitating their use as a valuable resource for clinical research.

By integrating data from these comprehensive datasets, this study capitalizes on their scale and diversity to allow a robust analysis of ICU readmission risks in patients with ICH. The use of MIMIC databases supports the investigation of clinically significant patterns and outcomes in a large, heterogeneous cohort of patients in the ICU, thus enhancing the generalizability and reliability of the findings.
\subsection{Study Population}\label{subsec:Study Population}
The study population consisted of patients diagnosed with ICH, identified using the International Classification of Diseases, Ninth Revision (ICD-9) code 431, and Tenth Revision (ICD-10) codes I610–I616 and I618–I619. Patients were included if ICH was recorded as a proposed or secondary diagnosis. Exclusion criteria were applied to ensure a well-defined population, omitting patients younger than 18 years, those with an ICU stay of less than 24 hours, and those who died during or immediately after their first ICU transfer. For patients with multiple ICU admissions, only data from the first ICU stay were included to maintain consistency and reduce redundancy.

After applying these criteria, 871 patients were identified from the MIMIC-III database and 1,445 from the MIMIC-IV database. These data sets were merged into a single unified cohort to facilitate analysis. The combined data set was divided into two subsets: 80\% for model training, and 20\% for testing. This distribution ensures a balanced allocation of data, allowing for rigorous model development and performance evaluation. The workflow of extracting the study population is shown in \textbf{Figure 1.}

\begin{figure}[H]
\centering
\includegraphics[width=1\textwidth]{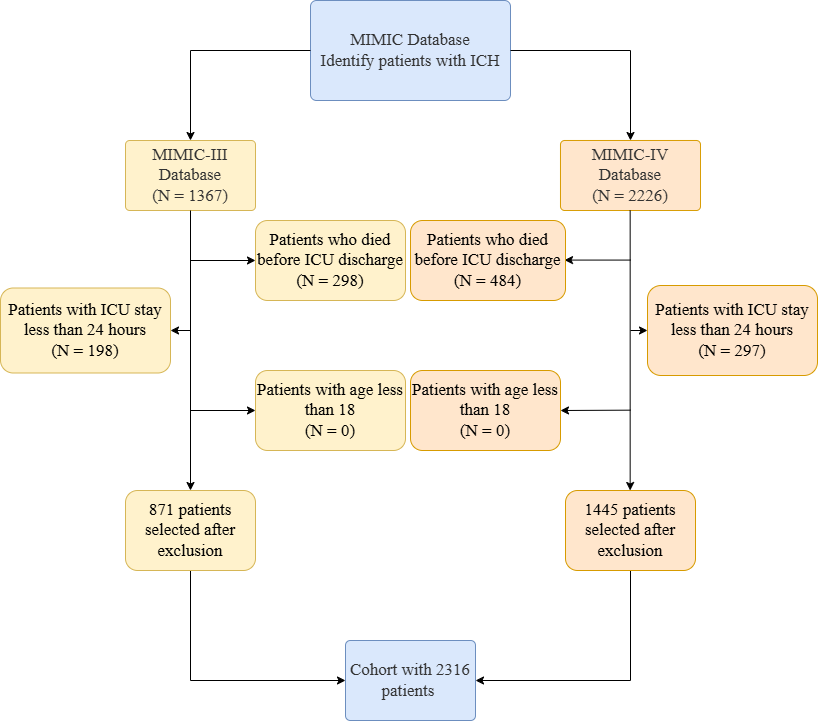}
\caption{Criterion of study population extraction}\label{population}
\end{figure}

\subsection{Feature Selection}\label{subsec:Identifying Features}
In our study, the feature selection process unfolded in multiple phases, including a comprehensive literature review, Recursive Feature Elimination(RFE) feature selection method, and experts' opinions. To start, we conducted an extensive review of the literature and considered input from experts to validate candidate features \cite{miao2024predicting,liotta2013predictors,chen2022machine,mercurio2024novel}. After this process, the dataset consisted of 33 features, representing various attributes relevant to the analysis. These features include demographics, vital signs, comorbidities, laboratory results, and medications.

We then applied a two-step feature selection process to reduce the dimensionality of the dataset. Given the potential for irrelevant or redundant features to negatively impact model performance, we first utilized RFE, followed by a refinement process incorporating clinical expertise.

In the first step, we performed RFE, a wrapper method that identifies the most relevant features based on their importance to the predictive model. RFE iteratively removes features with the least predictive value, optimizing the model performance at each stage \cite{sabouri2023machine,qiu2021hfs,ahmed2022integrated}. As a result, we identified a subset of 10 features that were deemed the most significant for predicting the target outcome. These features included: hospital stay, Alanine Aminotransferase (ALT), chloride, creatinine, sodium, monocytes, neutrophils, prothrombin time (PT), MCHC (Mean Corpuscular Hemoglobin Concentration), and International Normalized Ratio (INR).

Following this, we consulted with two clinical experts to incorporate their domain knowledge into the feature selection process. Based on their input, we added two additional features, age and SpO2, which were identified by the two experts as clinically relevant for the model. This expert-guided refinement resulted in a final set of 12 features: age, hospital stay, SpO2, ALT, chloride, creatinine, sodium, MCHC, monocytes, neutrophils, PT, and INR. The selected features are shown in \textbf{Table 1.}

This combination of statistical feature selection through RFE and expert-driven refinement ensured that the final feature set was both data-driven and clinically meaningful, contributing to a more robust and interpretable model.

\begin{table}[ht]
\centering
\caption{Selected clinical features}
\begin{tabular}{|c|l|}
\hline
\textbf{Category} & \textbf{Feature Name} \\
\hline
\textbf{Demographic Information} & Age \\
\hline
\textbf{Clinical Parameters}     & Hospital Stay \\
                                 & SpO2 \\
\hline
\textbf{Laboratory Measurements} & Alanine Aminotransferase (ALT) \\
                                 & Chloride \\
                                 & Creatinine \\
                                 & Sodium \\
                                 & MCHC (Mean Corpuscular Hemoglobin Concentration) \\
                                 & Monocytes \\
                                 & Neutrophils \\
                                 & PT (Prothrombin Time) \\
                                 & INR (International Normalized Ratio) \\
\hline
\end{tabular}
\end{table}

\begin{figure}[H]
\centering
\includegraphics[width=1.0\textwidth]{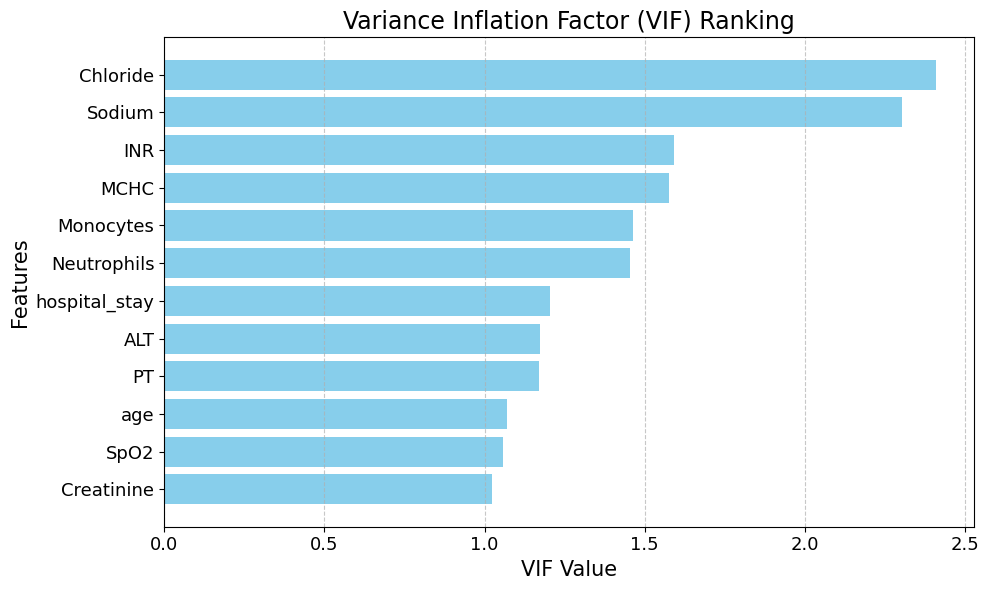}
\caption{Variance Inflation Factor of selected features.}
\label{fig:vif}
\end{figure}

The Variance Inflation Factor (VIF) was calculated to evaluate potential multicollinearity among predictors \cite{o2007caution}, as depicted in \textbf{Figure~\ref{fig:vif}}. All variables exhibited VIF values below the commonly accepted threshold of 5, indicating the absence of severe multicollinearity. The highest VIF was observed for "Chloride" \textbf{(2.41)} and "Sodium" \textbf{(2.30)}, which remain within acceptable limits. These results suggest that multicollinearity is unlikely to significantly impact the stability and interpretability of the regression model.

\subsection{Data Cleaning and Handling Missing Values}\label{subsec:Data Cleaning and Handling Missing Values}
The first step in preprocessing involved addressing missing values, handled based on feature type (categorical or numeric) and the proportion of missing data. 

For categorical features with 0-20\% missing values, we used SimpleImputer with the most frequent strategy, replacing missing values with the most common category. Categorical features with 20-100\% missing values were dropped to avoid introducing unnecessary noise.

For numeric features with 0-20\% missing values, KNN imputation was applied, predicting missing values based on neighboring data points. Numeric features with 20-50\% missing values were imputed using Iterative Imputation, which models each feature based on other features in the dataset.

\subsection{Statistical Analysis}\label{subsec:statistical Analysis}

To assess the differences between the training and test datasets, two statistic analyses were conducted \cite{demvsar2006statistical}. The first analysis compared the training and test datasets to evaluate their representativeness. The second analysis focused on comparing patients with readmission records to those without, aiming to identify potential factors associated with patient outcomes. Since all 12 features used in our final analysis were continuous variables, two-sided t-tests were exclusively used to assess differences in their means between the groups. A significance level of $P < 0.05$ was used as the threshold to determine statistical significance. This focused statistical approach ensured a precise evaluation of the datasets and provided insight into potential disparities that could influence subsequent analyses.

\subsection{Modeling}\label{subsec:Modeling}

To evaluate the performance of the proposed model, we applied a train-test split, allocating 80\% of the dataset for training and the remaining 20\% for testing. This approach ensures that the model is trained on a significant portion of the data while being evaluated on an independent set, allowing for reliable assessment of its generalization ability and minimizing the risk of overfitting.

The proposed model used in this study was an ANN model, selected for its ability to capture complex, non-linear relationships within the data. ANN are widely used in similar tasks due to their flexibility and capacity to model intricate, high-dimensional patterns that may be challenging to represent using traditional linear models. The model architecture was designed to incorporate ADASYN (Adaptive Synthetic Sampling), a technique that generates synthetic samples to address class imbalance in the dataset \cite{majhi2023wavelet,son2023improved,he2008adasyn}. ADASYN focuses on generating more samples in regions where the minority class is underrepresented, which helps improve the model's performance by preventing bias toward the majority class and enhancing the network's ability to classify underrepresented instances accurately.

The architecture of the neural network consisted of four hidden layers, along with an output layer consisting of a single neuron with a sigmoid activation function for binary classification tasks. Each hidden layer used the ReLU activation function, which is commonly employed in deep learning models for its ability to mitigate the vanishing gradient problem during training.

To optimize the model’s hyperparameters, we performed Grid Search \cite{yau2024machine,abad2020predicting,alsinglawi2024predicting}, which systematically tests different combinations of parameters to select the optimal configuration for the neural network, thereby improving its performance on the task.

To further benchmark the performance of our proposed model, we also implemented two widely-used machine learning algorithms as baseline models: XGBoost and Random Forest. These models were chosen due to their strong performance in classification tasks and their ability to handle complex, high-dimensional data effectively. Both models were trained using the same training and testing split, allowing for a fair comparison of their predictive accuracy and generalization ability relative to the proposed ANN model. The results from these baseline models provide valuable insights into the effectiveness of our approach in comparison to more traditional machine learning techniques.

\section{Results}\label{sec:Results}

\subsection{Statistical Comparison}\label{Cohort Comparison}

In the comparison of the training and test datasets, the t-tests revealed that all 12 characteristics had p-values less than 0.05, indicating statistically significant differences between the two datasets. In contrast, in the readmission and non-readmission comparison, the majority of variables showed statistically significant differences, with only three variables—hospital stay, ALT, and INR—exhibiting p-values greater than 0.05. This detailed statistical assessment provided important information on differences between datasets and highlighted key characteristics that may or may not be associated with survival outcomes. The comparison of population characteristics between the training and test datasets is presented in \textbf{Table 2}, while the comparison between the readmission and non-readmission datasets is shown in \textbf{Table 3}, with Group 0 representing the non-readmission dataset and Group 1 representing the readmission dataset.

\setlength{\tabcolsep}{10pt} 
\begin{table}[htbp]
\centering
\caption{Comparison of population features between training and test datasets.}
\begin{tabular}{lccc}
\toprule
Variable       & Training Mean (SD) & Test Mean (SD) & P-value \\
\midrule
Age            & 65.5 (15.5)     & 66.8 (14.8)    & 0.106 \\
Hospital Stay  & 13.3 (14.2)     & 13.1 (14.2)    & 0.771 \\
ALT            & 37.6 (124.5)    & 37.6 (63.2)    & 0.992 \\
Chloride       & 102.6 (4.9)     & 102.7 (4.6)    & 0.613 \\
Creatinine     & 1.0 (0.8)       & 1.0 (0.9)      & 0.434 \\
MCHC           & 33.3 (1.5)      & 33.2 (1.4)     & 0.088 \\
Monocytes      & 5.9 (3.0)       & 6.2 (4.1)      & 0.102 \\
Neutrophils    & 76.6 (9.9)      & 76.1 (10.2)    & 0.355 \\
PT             & 13.3 (3.0)      & 13.3 (3.0)     & 0.990 \\
SpO2           & 96.3 (3.2)      & 96.1 (3.7)     & 0.370 \\
INR            & 1.2 (0.2)       & 1.2 (0.3)      & 0.334 \\
\bottomrule
\end{tabular}
\label{tab:model_performance_comparison}
\end{table}

\setlength{\tabcolsep}{10pt} 
\begin{table}[htbp]
\centering
\caption{Comparison of population features between readmission and non-readmission datasets.}
\begin{tabular}{lccc}
\toprule
Variable & Group 0 & Group 1 & P-value \\
\midrule
age & 65.1 (15.5) & 73.5 (13.4) & 0.0000 \\
hospital stay & 13.4 (13.9) & 11.6 (20.5) & 0.4276 \\
ALT & 33.8 (39.8) & 118.6 (555.6) & 0.1684 \\
Chloride & 102.4 (4.6) & 106.8 (7.5) & 0.0000 \\
Creatinine & 0.9 (0.8) & 1.4 (1.7) & 0.0093 \\
Sodium & 138.7 (3.6) & 140.8 (6.5) & 0.0049 \\
MCHC & 33.3 (1.5) & 33.9 (1.4) & 0.0004 \\
Monocytes & 5.9 (3.0) & 3.9 (1.8) & 0.0000 \\
Neutrophils & 76.4 (9.6) & 82.3 (13.8) & 0.0002 \\
PT & 13.2 (3.0) & 14.3 (2.7) & 0.0010 \\
SpO2 & 96.5 (2.7) & 91.9 (7.5) & 0.0000 \\
INR & 1.2 (0.2) & 1.2 (0.3) & 0.1787 \\
\bottomrule
\end{tabular}
\label{tab:model_performance_comparison2}
\end{table}

\subsection{Model Performance}\label{Model Performance}

The final configuration of the ANN architecture consisted of four hidden layers, with 128 neurons in the first layer, 64 neurons in the second layer, 32 neurons in the third layer, and 16 neurons in the fourth layer. To prevent overfitting and enhance generalization, L2 regularization was applied to each layer, with regularization parameters (lambda) set to 0.03 for the first and second layers, 0.04 for the third layer, and 0.03 for the fourth layer. These configurations were identified as the optimal settings through the grid search process, resulting in an improved model performance in terms of predictive accuracy and robustness.

As detailed in \textbf{Table 4}, the proposed ANN model exhibited superior predictive performance compared to the baseline methods in the task of forecasting ICU readmissions for patients diagnosed with ICH.

In terms of performance metrics, the ANN model achieved an Area Under the Receiver Operating Characteristic Curve (AUROC) of \textbf{0.899 (95\% Confidence Interval: 0.860–0.911)}, the highest among the models tested. Additionally, the model achieved an accuracy of \textbf{0.881}, a sensitivity (recall) of \textbf{0.893}, and a specificity of \textbf{0.796}. These results underscore the model’s ability to accurately identify high-risk patients in need of close monitoring, while maintaining a balanced trade-off between sensitivity and specificity. The relatively high sensitivity indicates that the model is particularly effective at correctly identifying patients who are likely to be readmitted, while the specificity reflects its capability to correctly classify patients who are unlikely to require readmission. Collectively, these metrics highlight the robustness and reliability of the ANN model in capturing complex, non-linear relationships inherent in the data, demonstrating its potential as a powerful tool for predicting ICU readmissions in ICH patients.
The AUROC-curves for our proposed model and baseline models are shown in \textbf{Figure 3.}

\begin{figure}[htbp]
\centering
\includegraphics[width=0.9\textwidth]{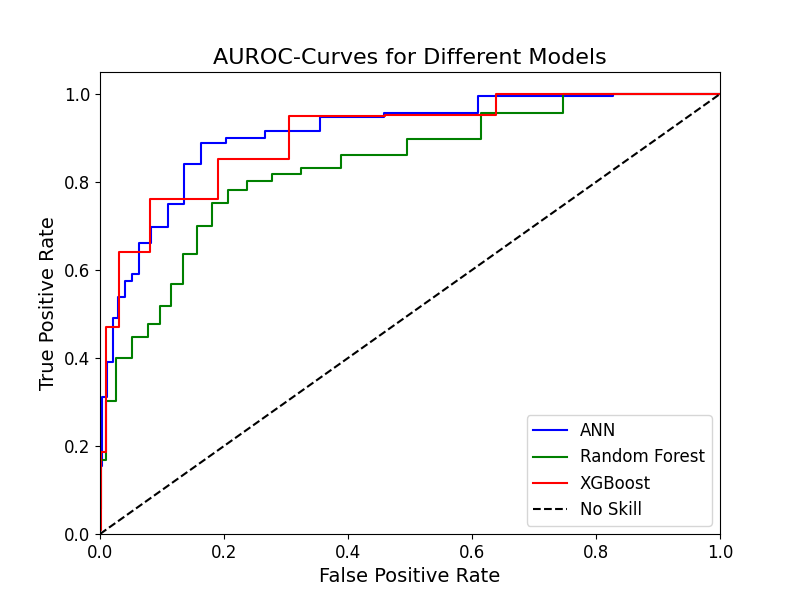}
\caption{AUROC-curves for test set of our three Machine Learning Models}
\label{fig:heatmap}
\end{figure}

\begin{table}[htbp]
\centering
\huge 
\renewcommand{\arraystretch}{1.2} 
\caption{Performance metrics of three Machine Learning models}
\label{tab:performance_metrics}
\resizebox{\columnwidth}{!}{%
\begin{tabular}{lccccc}
\toprule
\textbf{Best Method} & \textbf{Accuracy} & \textbf{AUROC} & \textbf{Sensitivity} & \textbf{Specificity} \\ \midrule
\textbf{ANN} & \textbf{0.881} & \textbf{0.899 (95\% CI: 0.860--0.911)}  & \textbf{0.893} & \textbf{0.796} \\ 
RandomForest        & 0.834 & 0.818 (95\% CI: 0.796--0.843) &  0.782 & 0.762 \\ 
XGBoost  & 0.807 & 0.870 (95\% CI: 0.846--0.881) &  0.853 & 0.696 \\ \bottomrule
\end{tabular}%
}
\end{table}

\subsection{Feature Importance Using SHAP}\label{Feature Importance Using SHAP}
To interpret the predictions made by our model, we employed Shapley Additive Explanations (SHAP), a powerful tool for understanding the contribution of individual features to the model output \cite{hu2022explainable,jiang2021explainable,gao2024prediction}. SHAP analysis was performed using the test dataset, revealing insightful patterns about the importance and directionality of characteristics in predicting ICU readmission for patients with ICH.

\begin{figure}[htbp]
    \centering
    \includegraphics[width=0.8\textwidth]{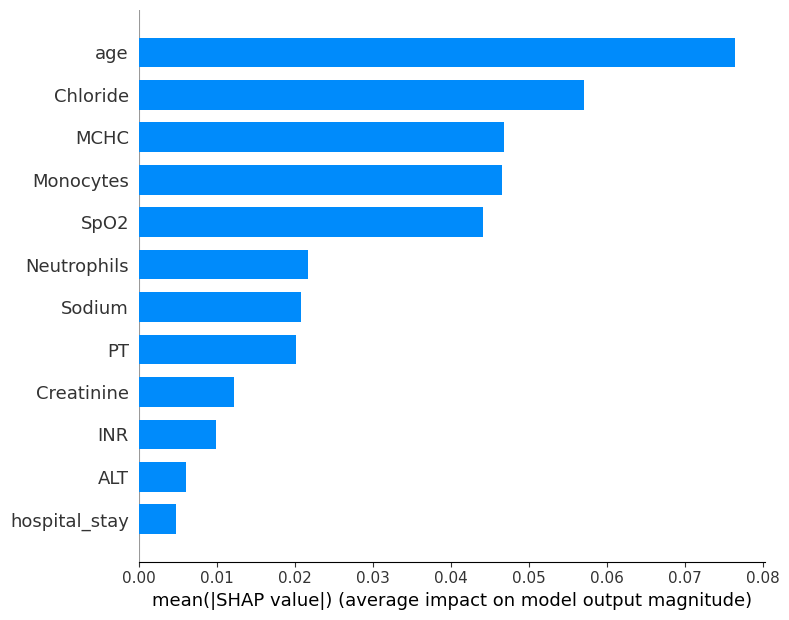}
    \caption{Feature importance ranking based on mean absolute SHAP values}
    \label{fig:shap_bar}
\end{figure}

\textbf{Figure~\ref{fig:shap_bar}} presents the ranking of the importance of features based on the mean absolute SHAP values. The results indicate that \textbf{age} is the most influential predictor, followed by \textbf{Chloride}, \textbf{MCHC}, and \textbf{Monocytes}. These features exhibited significant contributions to the model's output, highlighting their critical roles in predicting ICU readmissions. In particular, physiological parameters such as \textbf{SpO2}, \textbf{Neutrophils}, and \textbf{Sodium} also ranked highly, reflecting the relevance of clinical and demographic factors in the prediction process.

\begin{figure}[htbp]
    \centering
    \includegraphics[width=0.8\textwidth]{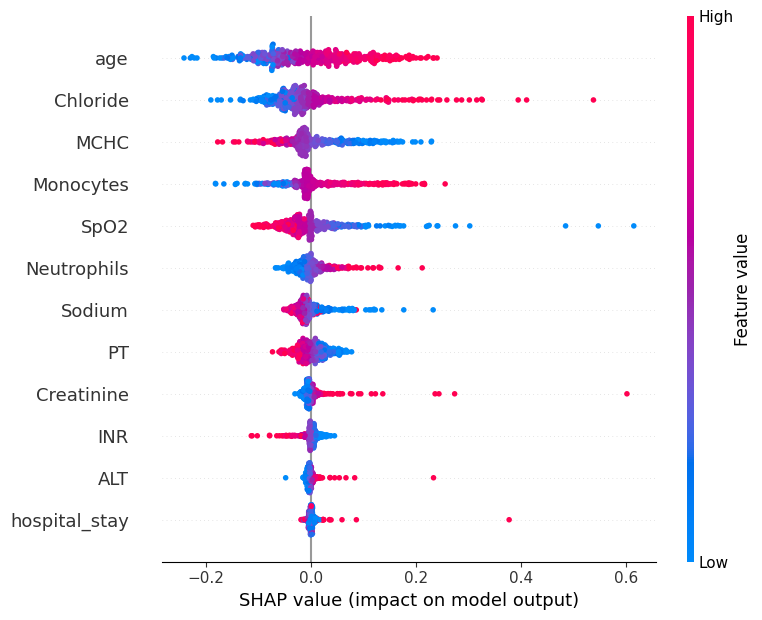}
    \caption{SHAP summary plot showing the distribution of SHAP values for each feature. Each dot represents an individual prediction, colored by the feature value (blue: low, red: high).}
    \label{fig:shap_summary}
\end{figure}

To further investigate how individual feature values influence predictions, \textbf{Figure~\ref{fig:shap_summary}} illustrates the SHAP summary plot, which shows the distribution of SHAP values for each feature. Each dot in the plot represents the data of a single patient, colored by the feature value (blue for low and red for high). For example, higher \textbf{age} values are associated with higher SHAP values, suggesting a higher likelihood of readmission. Similarly, abnormal levels of \textbf{Chloride}, \textbf{Monocytes}, and \textbf{SpO2} appear to significantly impact the predictions of the model, either increasing or decreasing the predicted risk.

These findings underscore the ability of our model to integrate diverse clinical and demographic data, leveraging the nonlinear interactions captured by the neural network. In addition, SHAP analysis improves the interpretability of the model, providing actionable insights for clinicians. For example, monitoring high-risk patients with advanced age, abnormal Chloride levels, or low SpO2 could guide early interventions and resource allocation in ICU settings.

\subsection{Comparison with Best Existing Models}\label{Comparison with Published Models}
In this study, we developed a predictive model using a Neural Network to predict ICU readmissions in ICH patients. Our results were compared against the best existing study~\cite{miao2024predicting}., which employed LightGBM as their proposed model for predicting ICU readmissions.  \textbf{Table~\ref{tab:comparison}} presents a detailed comparison between our work and the referred study.

\begin{table}[ht]
\centering
\huge
\renewcommand{\arraystretch}{1.2} 
\caption{Comparison of model performance metrics with referred work.}
\label{tab:comparison}
\resizebox{\columnwidth}{!}{%
\begin{tabular}{lcccccc}
\toprule
\textbf{Work Type} & \textbf{Best Method} & \textbf{Accuracy} & \textbf{AUROC}  & \textbf{Sensitivity} & \textbf{Specificity} \\ \midrule
\textbf{Ours}  & ANN & \textbf{0.881} & \textbf{0.899 (95\% CI: 0.860–0.911)} &  \textbf{0.893} & 0.796 \\
\textbf{Referred} & LightGBM    & 0.862          & 0.736 (95\% CI: 0.668--0.801)            & 0.226          & \textbf{0.943} \\ \bottomrule
\end{tabular}%
}
\end{table}

The comparison highlights the superior performance of our ANN model in key metrics. Specifically, our model achieved the highest \textbf{AUROC (0.899, 95\% CI: 0.860–0.911)}. The effectiveness of our model demonstrated an improvement of \textbf{4. 29\%} in AUROC compared to the study by Miao et al., who reported an AUROC of \textbf{0.736 (95\% CI: 0.668--0.801)}.

Our model achieved a recall of \textbf{0.893}, significantly exceeding the LightGBM model's recall of \textbf{0.226}. This result underscores the effectiveness of our approach in accurately identifying high-risk patients while maintaining a balanced trade-off between sensitivity and specificity.

In addition to superior performance in AUROC and sensitivity, our model provides a simpler architecture and a systematic feature selection process. By leveraging clinically relevant variables and advanced optimization techniques, we ensured the interpretability and efficiency of the model. The focus of the study by Miao et al. is on specificity (\textbf{0.943}) comes at the expense of recall, which limits its applicability for the early identification of high-risk cases.

Our results demonstrate a significant advancement in predicting ICU readmissions for ICH patients, offering better sensitivity and overall predictive performance. These findings support the potential of deep learning approaches, such as our ANN mode, in improving patient outcomes and optimizing resource allocation in critical care settings.

\section{Discussion}\label{sec:Discussion}
\subsection{Existing model compilation summary}\label{Model compilation summary}
In our study, we proposed an ANN model to predict ICU mortality in patients undergoing invasive mechanical ventilation. The result of our mortality prediction model was better than the best existing literature~\cite{miao2024predicting}, which highlights the superior effectiveness, robustness, and ability of our model in identifying high-risk patients compared to their approach.

Although the result of the existing literature effectively predicted mortality rates among ICU patients, it exhibited certain limitations. They used a total of 44 variables after feature selection to predict the outcome of the model. This approach may raise concerns related to model complexity and overfitting. In addition, the outcome of their research was considered unsatisfactory and inadequate for practical use in clinical exercise.

Furthermore, our study incorporates a rigorous assessment of multicollinearity using the VIF, ensuring that all selected features exhibit VIF values below the threshold of 5. This step, absent in the reference methodology, enhances the stability and reliability of our regression model by mitigating potential multicollinearity issues.

Furthermore, the Random Forest and XGBoost baseline models had AUROCs of 0.818 and 0.870, respectively, indicating that the selected features are highly predictive and relevant for the task of ICU readmission prediction. This suggests that the dataset and feature selection process effectively captured critical patterns and relationships, allowing even simpler models to achieve competitive results.

\subsection{limitation}\label{limitation}
Despite the promising results of our study, several limitations must be acknowledged. First, our analysis was conducted on data from a single source, the MIMIC-III and MIMIC-IV databases, which represents a specific cohort from a limited geographic region. As a result, the generalizability of our findings to other populations and healthcare systems may be restricted. Future studies should validate the model on external multicenter datasets to ensure its applicability in diverse clinical settings.

Second, although we addressed class imbalance using Random Oversampling and ADASYN, these techniques may introduce noise into the dataset, potentially affecting the reliability of the model. Further investigation into alternative methods for handling imbalance, such as ensemble techniques or novel synthetic sampling methods, could enhance the robustness of the model.

Third, while our ANN architecture demonstrated strong predictive performance, the complexity of deep learning models often limits interpretability. Although we mitigated this through systematic feature selection and SHAP analysis, simpler models like logistic regression or decision trees may be more suitable in settings where interpretability is critical.

Finally, our reliance on structured clinical data excludes potentially valuable information from unstructured sources such as free-text clinical notes, imaging data, or genetic profiles. Future work should explore integrating these data types to improve model performance and provide a more holistic view of patient health.

By addressing these limitations, future research can build on our findings to develop more generalizable, interpretable, and clinically impactful models to predict readmissions from the ICU in ICH patients.

\section{Conclusion}\label{sec:Conclusion}
This study developed a neural network model to predict ICU readmissions in ICH patients, demonstrating superior performance compared to baseline machine learning models. The results highlight the effectiveness of leveraging advanced neural networks for clinical prediction tasks.

Our systematic feature selection ensured that the model remained interpretable and clinically relevant, identifying key predictors that provide actionable insights for healthcare professionals. Robust preprocessing and optimization techniques further enhanced the model’s performance and generalizability. The lightweight and scalable architecture of the model makes it suitable for real-world clinical deployment, facilitating rapid and accurate predictions. 

Future research could validate this approach across diverse datasets and explore its applicability to other clinical outcomes. These findings underscore the potential of machine learning to improve decision-making and resource allocation in critical care settings.

%
%
\bibliography{reference}

\begin{thebibliography}{10}
\providecommand{\url}[1]{\texttt{#1}}
\providecommand{\urlprefix}{URL }

\bibitem{magid2022cerebral}
Magid-Bernstein, J., Girard, R., Polster, S., Srinath, A., Romanos, S., Awad, I.A., Sansing, L.H.: Cerebral hemorrhage: pathophysiology, treatment, and future directions. Circulation research  130(8),  1204--1229 (2022)

\bibitem{benjamin2019american}
Benjamin, E.J., Muntner, P., Alonso, A., Bittencourt, M.S., Callaway, C.W., Carson, A.P., Chamberlain, A.M., Chang, A.R., Cheng, S., Das, S.R., et~al.: American heart association council on epidemiology and prevention statistics committee and stroke statistics subcommittee. Heart disease and stroke statistics-2019 update: a report from the American Heart Association. Circulation  139(10),  e56--e528 (2019)

\bibitem{feigin2017global}
Feigin, V.L., Norrving, B., Mensah, G.A.: Global burden of stroke. Circulation research  120(3),  439--448 (2017)

\bibitem{van2010incidence}
Van~Asch, C.J., Luitse, M.J., Rinkel, G.J., van~der Tweel, I., Algra, A., Klijn, C.J.: Incidence, case fatality, and functional outcome of intracerebral haemorrhage over time, according to age, sex, and ethnic origin: a systematic review and meta-analysis. The Lancet Neurology  9(2),  167--176 (2010)

\bibitem{pinho2019intracerebral}
Pinho, J., Costa, A.S., Ara{\'u}jo, J.M., Amorim, J.M., Ferreira, C.: Intracerebral hemorrhage outcome: a comprehensive update. Journal of the neurological sciences  398,  54--66 (2019)

\bibitem{globalich2020}
Krishnamurthi, R., Ikeda, T., Feigin, V.: Global, regional and country-specific burden of ischaemic stroke, intracerebral haemorrhage and subarachnoid haemorrhage: A systematic analysis of the global burden of disease study 2017. Neuroepidemiology  54(2),  171--179 (02 2020), \url{https://doi.org/10.1159/000506396}

\bibitem{krishnamurthi2020global}
Krishnamurthi, R.V., Ikeda, T., Feigin, V.L.: Global, regional and country-specific burden of ischaemic stroke, intracerebral haemorrhage and subarachnoid haemorrhage: a systematic analysis of the global burden of disease study 2017. Neuroepidemiology  54(2),  171--179 (2020)

\bibitem{mcgurgan2021acute}
McGurgan, I.J., Ziai, W.C., Werring, D.J., Salman, R.A.S., Parry-Jones, A.R.: Acute intracerebral haemorrhage: diagnosis and management. Practical Neurology  21(2),  128--136 (2021)

\bibitem{kirshner2021management}
Kirshner, H., Schrag, M.: Management of intracerebral hemorrhage: update and future therapies. Current Neurology and Neuroscience Reports  21,  1--5 (2021)

\bibitem{elias2024new}
Elias, M., Robertson, N., Hughes, T.: New approaches in the management of intracranial haemorrhage. Journal of Neurology  271(9),  6393--6395 (2024)

\bibitem{hosein2014meta}
Hosein, F.S., Roberts, D.J., Turin, T.C., Zygun, D., Ghali, W.A., Stelfox, H.T.: A meta-analysis to derive literature-based benchmarks for readmission and hospital mortality after patient discharge from intensive care. Critical Care  18,  1--12 (2014)

\bibitem{maharaj2018utility}
Maharaj, R., Terblanche, M., Vlachos, S.: The utility of icu readmission as a quality indicator and the effect of selection. Critical Care Medicine  46(5),  749--756 (2018)

\bibitem{li2024machine}
Li, H., Ashrafi, N., Kang, C., Zhao, G., Chen, Y., Pishgar, M.: A machine learning-based prediction of hospital mortality in mechanically ventilated icu patients. Plos one  19(9),  e0309383 (2024)

\bibitem{esteva2019guide}
Esteva, A., Robicquet, A., Ramsundar, B., Kuleshov, V., DePristo, M., Chou, K., Cui, C., Corrado, G., Thrun, S., Dean, J.: A guide to deep learning in healthcare. Nature medicine  25(1),  24--29 (2019)

\bibitem{tsiknakis2021deep}
Tsiknakis, N., Theodoropoulos, D., Manikis, G., Ktistakis, E., Boutsora, O., Berto, A., Scarpa, F., Scarpa, A., Fotiadis, D.I., Marias, K.: Deep learning for diabetic retinopathy detection and classification based on fundus images: A review. Computers in biology and medicine  135,  104599 (2021)

\bibitem{rajkomar2018scalable}
Rajkomar, A., Oren, E., Chen, K., Dai, A.M., Hajaj, N., Hardt, M., Liu, P.J., Liu, X., Marcus, J., Sun, M., et~al.: Scalable and accurate deep learning with electronic health records. NPJ digital medicine  1(1),  1--10 (2018)

\bibitem{chen2016xgboost}
Chen, T., Guestrin, C.: Xgboost: A scalable tree boosting system. In: Proceedings of the 22nd acm sigkdd international conference on knowledge discovery and data mining. pp. 785--794 (2016)

\bibitem{liu2022dynamic}
Liu, S., Fu, B., Wang, W., Liu, M., Sun, X.: Dynamic sepsis prediction for intensive care unit patients using xgboost-based model with novel time-dependent features. IEEE Journal of Biomedical and Health Informatics  26(8),  4258--4269 (2022)

\bibitem{zhou2022construction}
Zhou, Y., Yu, Z., Liu, L., Wei, L., Zhao, L., Huang, L., Wang, L., Sun, S.: Construction and evaluation of an integrated predictive model for chronic kidney disease based on the random forest and artificial neural network approaches. Biochemical and biophysical research communications  603,  21--28 (2022)

\bibitem{johnson2016mimic}
Johnson, A.E., Pollard, T.J., Shen, L., Lehman, L.w.H., Feng, M., Ghassemi, M., Moody, B., Szolovits, P., Anthony~Celi, L., Mark, R.G.: Mimic-iii, a freely accessible critical care database. Scientific data  3(1),  1--9 (2016)

\bibitem{johnson2023mimic}
Johnson, A.E., Bulgarelli, L., Shen, L., Gayles, A., Shammout, A., Horng, S., Pollard, T.J., Hao, S., Moody, B., Gow, B., et~al.: Mimic-iv, a freely accessible electronic health record dataset. Scientific data  10(1), ~1 (2023)

\bibitem{miao2024predicting}
Miao, J., Zuo, C., Cao, H., Gu, Z., Huang, Y., Song, Y., Wang, F.: Predicting icu readmission risks in intracerebral hemorrhage patients: Insights from machine learning models using mimic databases. Journal of the Neurological Sciences  456,  122849 (2024)

\bibitem{liotta2013predictors}
Liotta, E.M., Singh, M., Kosteva, A.R., Beaumont, J.L., Guth, J.C., Bauer, R.M., Prabhakaran, S., Rosenberg, N.F., Maas, M.B., Naidech, A.M.: Predictors of 30-day readmission after intracerebral hemorrhage: a single-center approach for identifying potentially modifiable associations with readmission. Critical care medicine  41(12),  2762--2769 (2013)

\bibitem{chen2022machine}
Chen, T., Madanian, S., Airehrour, D., Cherrington, M.: Machine learning methods for hospital readmission prediction: systematic analysis of literature. Journal of Reliable Intelligent Environments  8(1),  49--66 (2022)

\bibitem{mercurio2024novel}
Mercurio, G., Gottardelli, B., Lenkowicz, J., Patarnello, S., Bellavia, S., Scala, I., Rizzo, P., de~Belvis, A.G., Del~Signore, A.B., Maviglia, R., et~al.: A novel risk score predicting 30-day hospital re-admission of patients with acute stroke by machine learning model. European Journal of Neurology  31(3),  e16153 (2024)

\bibitem{sabouri2023machine}
Sabouri, M., Rajabi, A.B., Hajianfar, G., Gharibi, O., Mohebi, M., Avval, A.H., Naderi, N., Shiri, I.: Machine learning based readmission and mortality prediction in heart failure patients. Scientific Reports  13(1),  18671 (2023)

\bibitem{qiu2021hfs}
Qiu, Y., Ding, S., Yao, N., Gu, D., Li, X.: Hfs-lightgbm: A machine learning model based on hybrid feature selection for classifying icu patient readmissions. Expert Systems  38(3),  e12658 (2021)

\bibitem{ahmed2022integrated}
Ahmed, A., Ashour, O., Ali, H., Firouz, M.: An integrated optimization and machine learning approach to predict the admission status of emergency patients. Expert Systems with Applications  202,  117314 (2022)

\bibitem{o2007caution}
O’brien, R.M.: A caution regarding rules of thumb for variance inflation factors. Quality \& quantity  41,  673--690 (2007)

\bibitem{demvsar2006statistical}
Dem{\v{s}}ar, J.: Statistical comparisons of classifiers over multiple data sets. The Journal of Machine learning research  7,  1--30 (2006)

\bibitem{majhi2023wavelet}
Majhi, B., Kashyap, A.: Wavelet based ensemble models for early mortality prediction using imbalance icu big data. Smart Health  28,  100374 (2023)

\bibitem{son2023improved}
Son, B., Myung, J., Shin, Y., Kim, S., Kim, S.H., Chung, J.M., Noh, J., Cho, J., Chung, H.S.: Improved patient mortality predictions in emergency departments with deep learning data-synthesis and ensemble models. Scientific reports  13(1),  15031 (2023)

\bibitem{he2008adasyn}
He, H., Bai, Y., Garcia, E.A., Li, S.: Adasyn: Adaptive synthetic sampling approach for imbalanced learning. In: 2008 IEEE international joint conference on neural networks (IEEE world congress on computational intelligence). pp. 1322--1328. Ieee (2008)

\bibitem{yau2024machine}
Yau, F.F.F., Chiu, I.M., Wu, K.H., Cheng, C.Y., Lee, W.C., Chen, H.C., Cheng, C.I., Chen, T.Y.: Machine learning-based prediction of coronary care unit readmission: A multihospital validation study. Digital Health  10,  20552076241277030 (2024)

\bibitem{abad2020predicting}
Abad, Z.S.H., Maslove, D.M., Lee, J.: Predicting discharge destination of critically ill patients using machine learning. IEEE journal of biomedical and health informatics  25(3),  827--837 (2020)

\bibitem{alsinglawi2024predicting}
Alsinglawi, B.S., Alnajjar, F., Alorjani, M.S., Alshari, O., Novoa, M., Mubin, O.: Predicting hospital stay length using explainable machine learning. IEEE Access  (2024)

\bibitem{hu2022explainable}
Hu, C., Li, L., Li, Y., Wang, F., Hu, B., Peng, Z.: Explainable machine-learning model for prediction of in-hospital mortality in septic patients requiring intensive care unit readmission. Infectious Diseases and Therapy  11(4),  1695--1713 (2022)

\bibitem{jiang2021explainable}
Jiang, Z., Bo, L., Xu, Z., Song, Y., Wang, J., Wen, P., Wan, X., Yang, T., Deng, X., Bian, J.: An explainable machine learning algorithm for risk factor analysis of in-hospital mortality in sepsis survivors with icu readmission. Computer Methods and Programs in Biomedicine  204,  106040 (2021)

\bibitem{gao2024prediction}
Gao, J., Lu, Y., Ashrafi, N., Domingo, I., Alaei, K., Pishgar, M.: Prediction of sepsis mortality in icu patients using machine learning methods. BMC Medical Informatics and Decision Making  24(1),  228 (2024)

\end{thebibliography}

\end{document}